\documentclass[letterpaper, 10 pt, conference]{IEEEtran}
\IEEEoverridecommandlockouts

\usepackage{cite}
\usepackage{amsmath,amssymb,amsfonts}
\usepackage{algorithmic}
\usepackage{graphicx}
\usepackage{textcomp}
\usepackage{xcolor}
\usepackage{subcaption}
\usepackage{tikz}
\usepackage{pgfplots}
\usepackage{verbatim}
\usepackage{rotating}
\usepackage{natbib}
\usepackage{colortbl}
\usepackage{hhline}
\usepackage{multirow}

\setcitestyle{square}

\setcitestyle{numbers}
\usepackage{hyperref}
\renewcommand{\thetable}{\Roman{table}}
\usepackage{graphicx}
\usepackage{geometry}
\def\BibTeX{{\rm B\kern-.05em{\sc i\kern-.025em b}\kern-.08em
    T\kern-.1667em\lower.7ex\hbox{E}\kern-.125emX}}
\begin{document}

\title{Enhancing Accuracy and Robustness of Steering Angle Prediction with Attention Mechanism\\
}

\author{\IEEEauthorblockN{Swetha Nadella$^{*}$, Pramiti Barua$^{*}$, Jeremy C. Hagler, David J. Lamb, and Qing Tian }
\IEEEauthorblockA{\textit{Department of Computer Science} \\
\textit{Bowling Green State University}\\
Bowling Green, Ohio, USA \\
emails: \{nswetha, pbarua, jhagler, djlamb, qtian\}@bgsu.edu}\\


}

\maketitle

\begin{abstract}
In this paper, our focus is on enhancing steering angle prediction for autonomous driving tasks. We initiate our exploration by investigating two veins of widely adopted deep neural architectures, namely ResNets and InceptionNets. Within both families, we systematically evaluate various model sizes to understand their impact on performance. Notably, our key contribution lies in the incorporation of an attention mechanism to augment steering angle prediction accuracy and robustness. By introducing attention, our models gain the ability to selectively focus on crucial regions within the input data, leading to improved predictive outcomes. Our findings showcase that our attention-enhanced models not only achieve state-of-the-art results in terms of steering angle Mean Squared Error (MSE) but also exhibit enhanced adversarial robustness, addressing critical concerns in real-world deployment. For example, in our experiments on the Kaggle SAP and our created publicly available datasets, attention can lead to over 6\% error reduction in steering angle prediction and boost model robustness by up to 56.09\%.
\end{abstract}
\def\thefootnote{*}\footnotetext{These authors contributed equally.}

\section{Introduction}
As we approach the era of autonomous vehicles, the precise and reliable prediction of steering angles is essential in ensuring safe and efficient self-driving operations of a vehicle. The accuracy of this steering angle prediction directly impacts the vehicle's ability to navigate complex and dynamic environments. Deep learning models have emerged as powerful tools to tackle this challenge \citep{Navarro2021}. 
Although deep learning models can increase the accuracy of steering angle prediction, it is important to acknowledge their shortcomings, including inefficiency and susceptibility to adversarial attacks. Such issues prevent deep neural networks' wide deployment in real-world steering angle prediction.
This paper first embarks on a comprehensive exploration of the two most popular veins of deep neural networks (i.e., InceptionNet and ResNet variations) for the task of steering angle prediction. Specifically, we explore the two types of networks positioned towards the compact end of the complexity spectrum.
More importantly, in this paper, we introduce an attention mechanism to the task of steering angle prediction, which can enhance both the prediction accuracy and robustness.
For example, for ResNet32, the introduction of attention leads to a 6.83\% error reduction in steering angle prediction and a robustness increase of up to 53.96\%. Qualitatively, we demonstrate that through our incorporation of the attention mechanism, our models gain the capability to selectively concentrate on vital regions within the input images.

\begin{figure*}[!ht]
\centering{\includegraphics[width=0.80\linewidth]{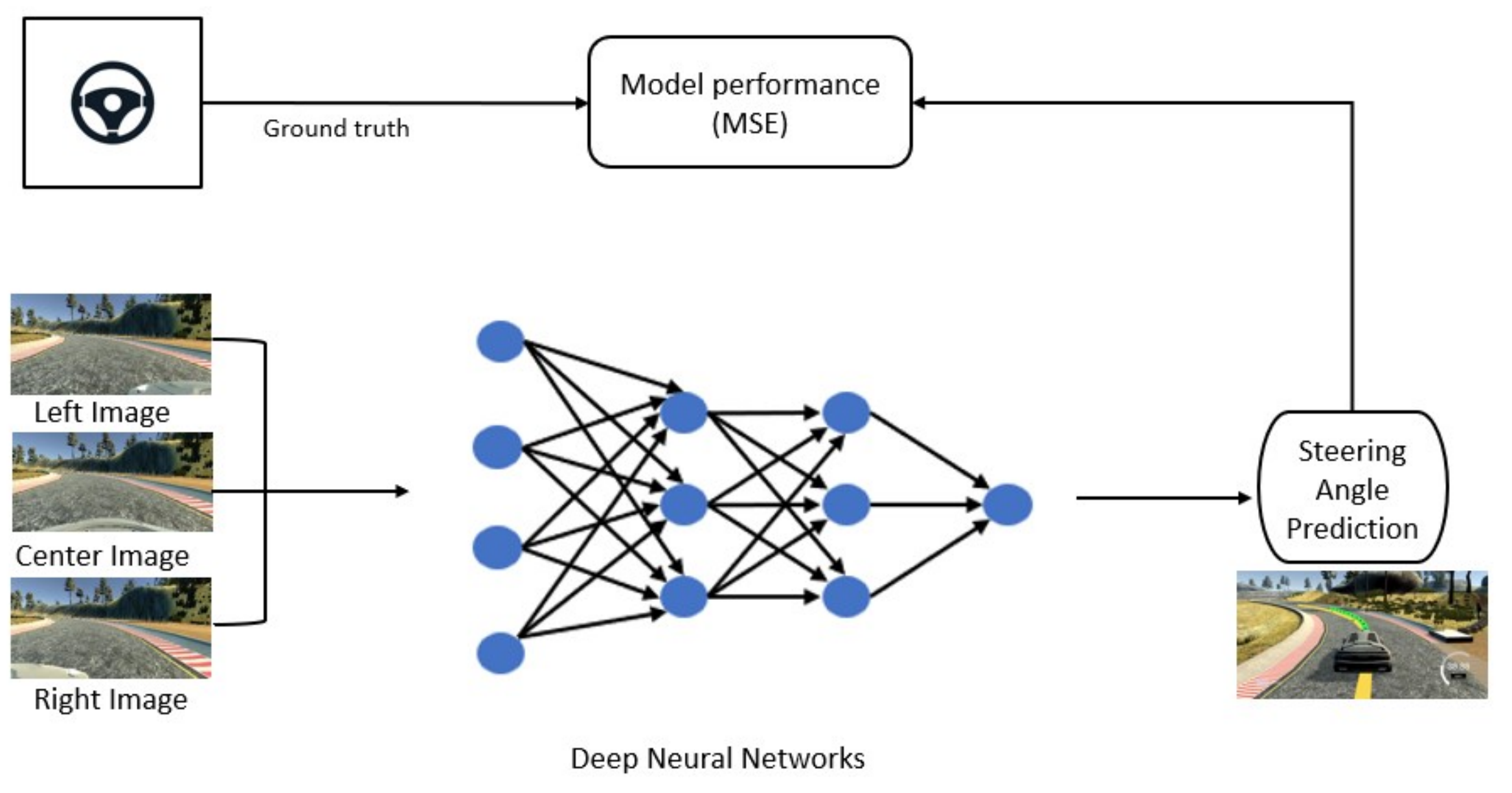}}
\vspace{0em}
\caption{Illustration of deep-learning-based steering angle prediction on visual data. The deep model is designed to analyze visual data captured by the left, central, and right cameras, facilitating the inference of steering angles.}
\label{flow}
\end{figure*}

\section{Related Work}

Autonomous driving is a rapidly evolving field that can potentially revolutionize the future of mobility.
The current state of the work done in steering angle prediction is early compared to other areas like visual recognition and detection. Steering angle prediction continues to rely more heavily on visual processing and understanding. Currently, even Tesla has begun to use pure vision in their cars \citep{Klender2021}. \citep{Gidado2020} survey different deep learning (DL) methods for steering angle prediction.
Steering angle prediction needs to be accurate, efficient, and robust. Most works in this direction focus on improving the accuracy with large models, but few discuss the efficiency and robustness issues.
While there are continuous changes in the field, the ResNet architecture remains a key player in steering angle prediction \cite{Oussama2019,Islam2021,Munir2022}.
For example, \citep{Oussama2019} surveyed different methods for achieving steering angle prediction, with most of the approaches relying on cameras, sensors, and/or radar. 
The results of this research compilation showed that a ResNet50-based architecture gives a better prediction of the due wheel angle. 
However, for on-board steering angle prediction, the cost to train and deploy ResNet50 is high, due to its large number of parameters \citep{Du2019}. 
\citep{Navarro2021} presents complex models for pushing the boundaries of autonomous driving. 
They show that these new end-to-end models have better optimization from the use of additional data information, like speed.
\citep{Qizwini2017} found InceptionNet as a best model for various autonomous driving tasks. \citep{White2020} discovered that InceptionNet can extract qualitatively equivalent information as its ResNet counterpart. 
Based on one single image, \citep{he2018aggregated} utilized an aggregated set of large VGG16 models to extract features for steering angle prediction. The LSTMs utilized also contributed to the approach's inefficiency.
On an unpublished dataset, \citep{zhu2020autonomous} proposed a multi-task MapNet to simultaneously search the passable area, detect the intersection, and predict the steering angle with many expensive 5$\times$5 conv filters.

The attention mechanism has been used in a variety of computer vision tasks. \citep{Wang2017} proposed a residual attention network 
to focus on an image’s most informative parts, allowing for more efficient and accurate classification. The attention mechanism works by adaptively weighing different features of the image, based on the context and spatial location. The attention mechanism includes the development of frameworks that incorporate both bottom-up and top-down structures in order to extract global features. At lower resolutions, the bottom-up feed-forward structure generates feature maps with robust semantic information, whereas the top-down network generates dense features for inference on each pixel. The present state-of-the-art performance in image segmentation has been attained by inserting skip connections between the bottom and top feature maps \citep{Long2014}. The study demonstrated the effectiveness of the attention mechanism in improving the network’s performance. To the best of our knowledge, only a few works have explored attention mechanisms for steering angle prediction, most of them are based on cumbersome backbones and a single input image \citep{he2018aggregated}.

The robustness of steering angle predictors is also of paramount importance. Prior studies have highlighted the impact of adversarial attacks in the realm of image classification \cite{Goodfellow2014}. 
This inherent susceptibility raises concerns about the safety and reliability of neural networks, particularly in the safety-critical field of self-driving cars \citep{Deng2020}. 
While adversarial attacks on neural networks have been extensively explored in different fields, mainly in image classification, the investigation of these attacks on autonomous steering angle prediction is limited, if any. 
In this paper, particularly in the context of steering angle prediction, we evaluate the vulnerability of neural networks to adversarial attacks such as Fast Gradient Sign Method (FGSM) \citep{Goodfellow2014} and Projected Gradient Descent (PGD) \citep{madry2017}. We can successfully improve the robustness of steering angle prediction by incorporating an attention mechanism.

\section{Methodology}
Figure \ref{flow} offers a high-level overview of our utilization of deep learning models and visual data for steering angle prediction.
Our deep models are created to process visual information from the left, central, and right cameras, enabling the inference of steering angles.
Given the limited resources for on-board steering angle prediction, we focus our attention on compact deep models in this paper. 
We will begin by providing a comparison of the two classic families of deep models for steering angle prediction. Subsequently, we will introduce an attention mechanism to enhance the precision and robustness of steering angle prediction models. 

\subsection{Classic Deep Models for Steering Angle Prediction}

Our first goal is to find a compact model with an accuracy score that is comparable to or better than its larger counterparts. This paper explores the two most popular veins of CNNs (ResNet \citep{He2016} and InceptionNet \citep{Szegedy2015} variations) with compact capacities.
We present the outcomes of various sizes for each of the two families. For example, we built and trained eight ResNet architectures; ResNet20, ResNet22, ResNet24, ResNet26, ResNet28, ResNet30, ResNet32, and ResNet34. We built seven InceptionNet models to match similar numbers of parameters so that we could compare the architectures at different complexity levels. The number of parameters and detailed configurations for each of the ResNet and InceptionNet models are illustrated in Table \ref{table:1} and \ref{table:2}. 
We will provide the comparison results in Sec. \ref{sec:experiments}.


\begin{table}[h!]
\centering
\caption{ResNet Model Parameters \& Block Layers}
\small
\begin{tabular}{c c c} 
 \hline
 ResNet Model & Parameters & Block Layers\\ [0.5ex] 
 \hline\hline
 ResNet20 & 12.8 million & (2,2,3,2)\\ 
 ResNet22 & 13.1 million & (2,3,3,2)\\
 ResNet24 & 14.3 million & (2,3,4,2)\\
 ResNet26 & 17.9 million & (3,3,3,3)\\
 ResNet28 & 19.1 million & (3,3,4,3)\\
 ResNet30 & 19.4 million & (3,4,4,3)\\
 ResNet32 & 20.6 million & (3,4,5,3)\\ 
 ResNet34 & 21.7 million & (3,4,6,3)\\  
 \hline
\end{tabular}
\label{table:1}
\end{table}
\begin{table}[h!]
\centering
\caption{InceptionNet Model Parameters \& Block Layers}
\small
\begin{tabular}{c c c} 
 \hline
 InceptionNet Model & Parameters & Block Layers \\ [0.5ex] 
 \hline\hline
 InceptionNet  & 13.0 million & (2,5,2)\\ 
 InceptionNet a & 14.5 million & (5,5,2) \\
 InceptionNet b & 15.5 million & (2,8,2)\\
 InceptionNet c & 17.0 million & (5,8,2)\\
 InceptionNet d & 17.6 million & (2,5,5)\\
 InceptionNet e & 20.2 million & (2,8,5)\\
 InceptionNet f & 21.7 million & (5,8,5)\\ 
 
 \hline
\end{tabular}
\label{table:2}
\end{table}


\subsection{Attention-aware Deep Steering Angle Prediction}\label{sec:attentionmodule}

\begin{figure*}
\centering
\includegraphics[width=0.9\linewidth, trim={0 0 0 0},clip]{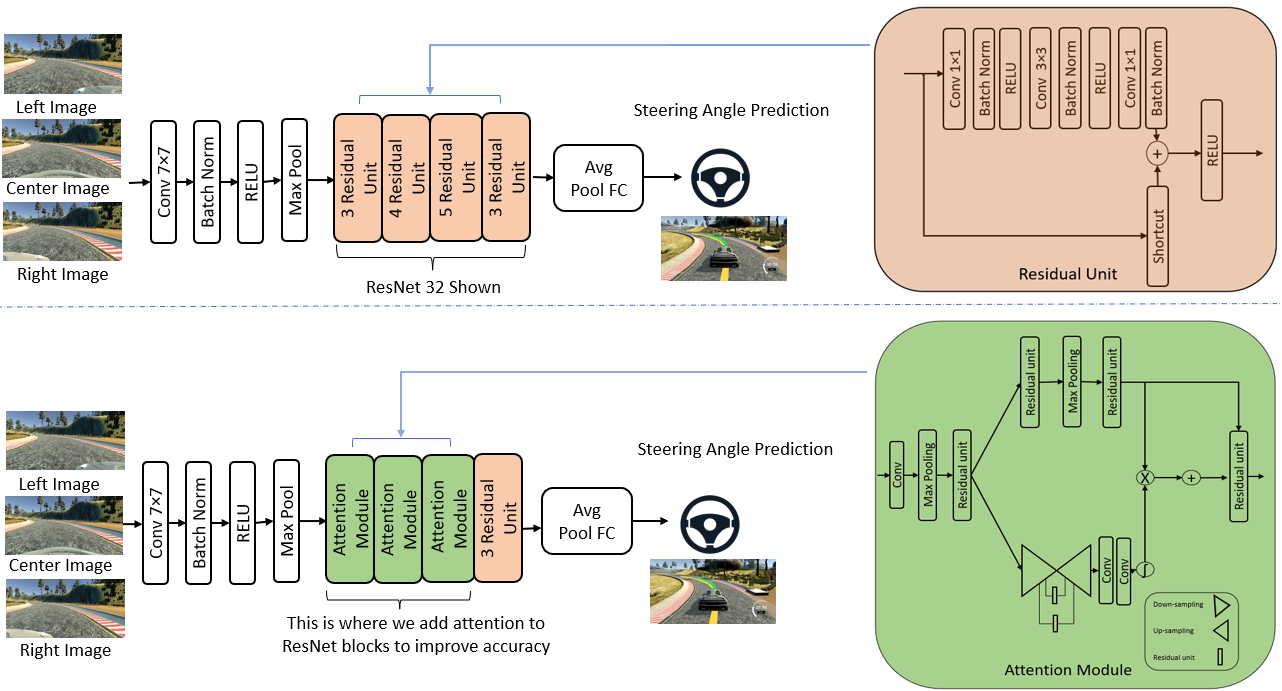}
\captionof{figure}{Comparison of ResNet32 and ResNet32 with attention (two models are of similar sizes). The attention module consists of two basic branches: the trunk branch and the attention mask branch. More details in Sec. \ref{sec:attentionmodule}.}
\label{flow_new}
\end{figure*}

In this section, we draw inspiration from the success of attention mechanisms \citep{Wang2017} and propose to add an attention mechanism to the task of steering angle prediction. As an example, we take the best-performing ResNet32 as the baseline model for our experiments. Figure \ref{flow_new} shows a comparison of our basic ResNet32 network and ResNet32 network with the addition of attention modules, where both networks have the same number of residual units to ensure a fair comparison.
As shown in Figure \ref{flow_new}, the attention module consists of two basic branches: the trunk branch and the attention mask branch. The trunk branch works as a standard encoder to extract features from the input. On the other hand, the attention mask branch learns a mask of the same size as the extracted features from the encoder on which the implementation of an effective attention mechanism relies. To extract the global features, the input features undergo several down-samplings. These global features are then extended to their original sizes through symmetrical up-sampling to obtain the mask. By calculating the dot product of the mask and the output of the trunk branch, the mask then escorts output features at each point.

The attention module includes three attention blocks, each of which consists of three operations: the feature fusion operation, the feature selection operation, and the feature amplification operation. The feature selection operation separates the useful features from the unimportant ones while the feature fusion operation combines the feature maps from the trunk network and the preceding attention block. The operation of feature amplification amplifies the crucial features to improve their representation. 

\subsection{Model Robustness Assessment}

The attention module, as explained previously, incorporates trunk and attention mask branches, facilitating an effective attention mechanism that selectively emphasizes essential features in the input data. In addition to enhancing accuracy through attention mechanisms, our research also delves into the models' ability to resist adversarial attacks, such as Fast Gradient Sign Method (FGSM) and Projected Gradient Descent (PGD). We evaluate how adversarial attacks impact our best-performing models, both with and without the attention mechanism. 
Through these adversarial tests, we aim to assess whether the attention mechanism enhances the model's security and its capacity to cope with adversarial manipulations \citep{Goodfellow2014,madry2017}.
To ensure fair comparisons, each network is configured with the same number of residual units.

\section{Experiments and Results}\label{sec:experiments}

\subsection{Environmental Setup}
Experiments were run on the Ohio Supercomputer Center's Pitzer cluster with NVIDIA Tesla V100 GPUs \citep{supercomp1987}. The deep learning framework used was PyTorch.
We use mean squared error (MSE) as the loss function:
\begin{equation}
MSE= \frac{1}{n} \displaystyle\sum_{n=1} ^{n} (r_{i} - \hat{r_{i}})^2\,,
\end{equation}
\noindent where $r_{i}$ is actual steering angle (in radians) and $\hat{r_{i}}$ is predicted steering angle (in radians). 
Models were trained for 50 epochs and the batch size tested for all models varied from 16 to 128 in search of the best outcome. Generally, the batch sizes of 32 were found to have the best results.

\subsection{Quantitative Results}
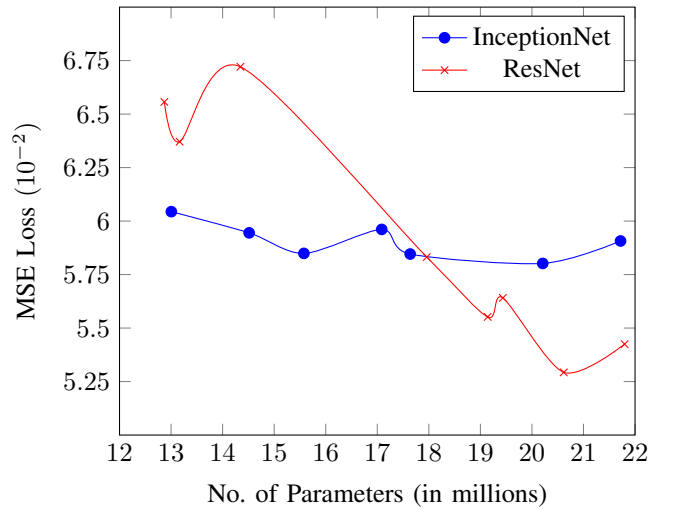
\begin{figure}[!htb]
\begin{center}
\scalebox{1} {
\begin{tikzpicture}
    \begin{axis}[
        xlabel=No. of Parameters (in millions),
        ylabel=MSE Loss $(10^{-2})$,
        xmin=12, xmax=22,
        ymin=5.0, ymax=7.0,
        xtick={12,13,14,15,16,17,18, 19, 20,21,22},
        ytick={ 5.25,5.5,5.75, 6.0,6.25, 6.5, 6.75}
        ]
    
    \addplot[smooth,mark=*,blue] plot coordinates {
        (13.004888,6.044)
        (14.514392,5.945)
        (15.576664, 5.849)
        (17.086168, 5.961)
        (17.638008, 5.846)
        (20.209784, 5.802)
        (21.719288, 5.907)
    };
    \addlegendentry{InceptionNet}

    \addplot[smooth,color=red,mark=x]
        plot coordinates {
            (12.870184,6.557)
            (13.165608,6.371)
            (14.346280,6.721)
            (17.960232, 5.832)
            (19.140904, 5.552)
            (19.436328, 5.641)
            (20.617000, 5.293)
            (21.797672, 5.425)
        };
    \addlegendentry{ResNet}
    \end{axis}
    \end{tikzpicture}}
\end{center}
    \vspace{-1em}
    \caption{ResNets vs. InceptionNets in terms of MSE loss vs number of parameters on the Kaggle SAP dataset.}
    \label{figure2}
\end{figure}

Our first analysis was with a publicly available Kaggle data set of approximately 97,330 images \citep{Kaggle2019}. In this paper, we refer to this dataset as the Kaggle Steering Angle Prediction dataset or simply the Kaggle SAP dataset. According to Table \ref{table3}, the ResNet32 model achieves the lowest error of 0.05293.
This is a competitive score as our ResNet model performed similarly (or better) than that of recent works \cite{ijaz2021, Oinar2022}.
The InceptionNet architecture performed worse and had the lowest score of 0.05846. 
In our experiments, the training loss of ResNet models also drops faster than Inception counterparts.
Figure \ref{figure2} plots all of the tested models with their complexity measure on the x-axis and their MSE scores on the y-axis. The plotted lines intersect around 17.9 or 18 million parameters.
\begin{table}[h!]
\centering
\caption{MSE Scores on the Kaggle SAP dataset for series of ResNet and Inception architectures.}
\small
\begin{tabular}{c c c c } 
 \hline
 ResNet Model & MSE & InceptionNet Model & MSE\\ [0.5ex] 
 \hline\hline
 ResNet20  & 0.06557 & InceptionNet & 0.06044\\ 
 ResNet22  & 0.06371 & InceptionNet a & 0.05945\\
 ResNet24  & 0.06721 & InceptionNet b & 0.05849\\
 ResNet26 & 0.05832 & \textbf{InceptionNet c} & \textbf{0.05846} \\
 ResNet28 & 0.05552 & InceptionNet d & 0.05961\\
 ResNet30 & 0.05641 & InceptionNet e & 0.05802\\
 \textbf{ResNet32} & \textbf{0.05293} & InceptionNet f & 0.05907\\ 
 ResNet34 & 0.05425 &&\\  
 \hline
\end{tabular}

\label{table3}
\end{table}

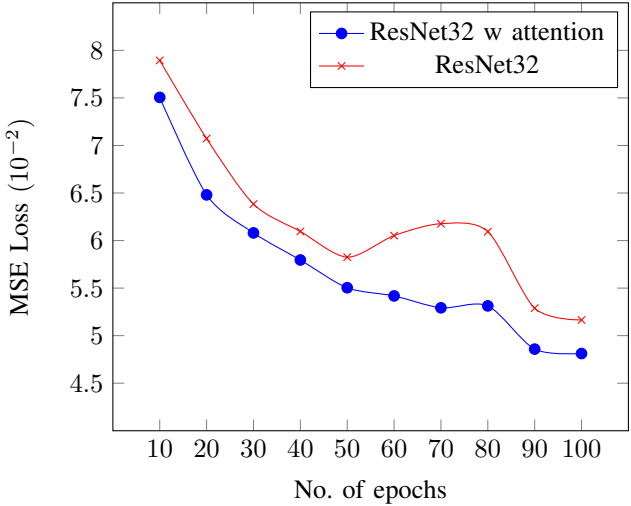
\begin{figure}
\begin{center}
\hspace{-0.5in}
\scalebox{1} {
\begin{tikzpicture}
    \begin{axis}[
        xlabel=No. of epochs,
        ylabel=MSE Loss $(10^{-2})$,
        xmin=0, xmax=110,
        ymin=4, ymax=8.5,
        xtick={10,20,30,40,50,60,70, 80, 90, 100},
        ytick={4.50,5.00, 5.50, 6.00, 6.50, 7.00, 7.50, 8.00}
        ]
    \addplot[smooth,mark=*,blue] plot coordinates {
        (10,7.5056907)
        (20,6.4803872)
        (30,6.0805413)
        (40,5.7955205)
        (50,5.5037914)
        (60,5.4176859)
        (70,5.2928171)
        (80,5.3129615)
        (90,4.8582893)
        (100,4.8118659)
    };
    \addlegendentry{ResNet32 w attention}

    \addplot[smooth,color=red,mark=x]
        plot coordinates {
            (10,7.8942212)
            (20,7.0727005)
            (30,6.3836283)
            (40,6.0956478)
            (50,5.8252398)
            (60,6.0522372)
            (70,6.1771978)
            (80,6.0938124)
            (90,5.2891241)
            (100,5.1648843)
        };
    \addlegendentry{ResNet32}
    \end{axis}
    \end{tikzpicture}}
\end{center}
\vspace{-1em}
\caption{MSE comparison between ResNet32 vs ResNet32 with attention modules on the Kaggle SAP dataset.}
    \label{mse-res32}
\end{figure}

\begin{figure}
    
\begin{center}
\hspace{-0.5in}
\scalebox{1} {
\begin{tikzpicture}
    \begin{axis}[
        xlabel=No. of epochs,
        ylabel=MSE Loss $(10^{-2})$,
        xmin=0, xmax=110,
        ymin=3.5, ymax=5.8,
        xtick={10,20,30,40,50,60,70,80,90,100},
        ytick={4.20, 4.60,5.00, 5.40, 5.80}
        ]
    \addplot[smooth,mark=*,blue] plot coordinates {
        (10,4.6298954)
        (20,4.3369272)
        (30,4.4884232)
        (40,4.3949584)
        (50,4.2272017)
        (60,4.2382651)
        (70,4.3373045)
        (80,4.1406841)
        (90,4.0761644)
        (100,4.0053548)
    };
    \addlegendentry{ResNet26 w attention}

    \addplot[smooth,color=red,mark=x]
        plot coordinates {
            (10,4.9390083)
            (20,5.3060975)
            (30,5.1836453)
            (40,4.6840318)
            (50,4.6384332)
            (60,4.6540305)
            (70,4.3620725)
            (80,4.4712992)
            (90,4.1939236)
            (100,4.2653428)
        };
    \addlegendentry{ResNet26}
    \end{axis}
    \end{tikzpicture}}
\end{center}
\vspace{-1em}
\caption{MSE comparison between ResNet26 vs ResNet26 with attention modules on the Custom dataset.}
    \label{fig:MSE-res26}
\end{figure}
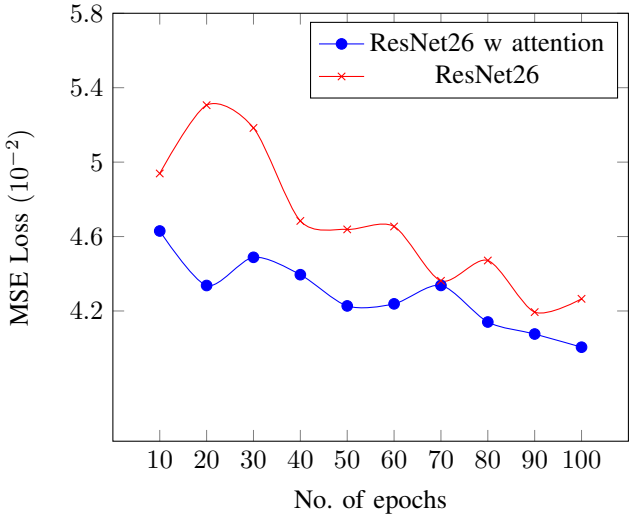
\begin {figure}
\begin{center}
\includegraphics[height=0.6\linewidth]{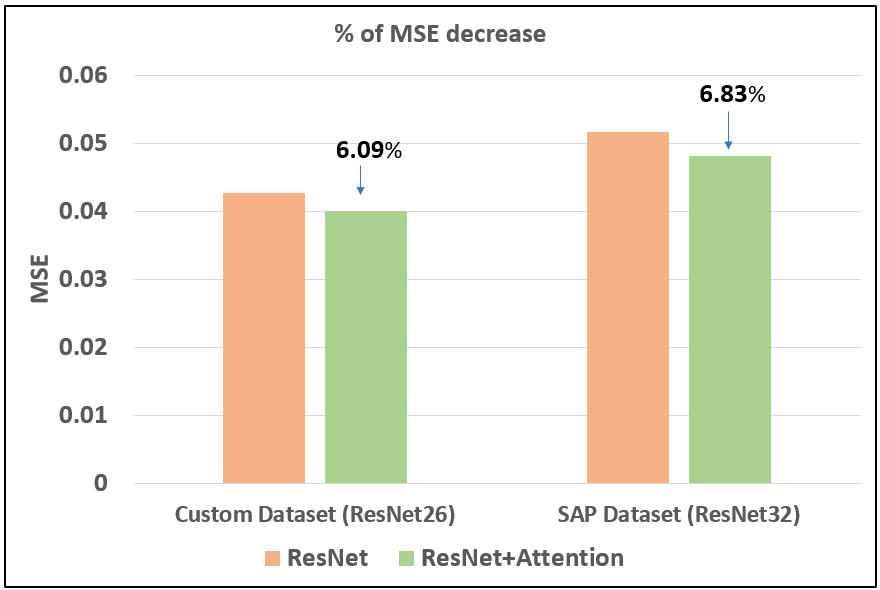}
\captionof{figure}{Performance improvement of ResNet-26 and ResNet-32 with attention modules added.}
\label{MSE-decrease}
\end{center}
\end {figure} 

To demonstrate the efficacy of our attention mechanism, we conducted a comparison between the baseline ResNet32 model and ResNet32 with attention modules, utilizing the Kaggle SAP dataset. The results depicted in Figure \ref{mse-res32} reveal a significant reduction in MSE during validation after incorporating attention modules into the ResNet32 model at around 70 epochs during training. The validation loss of our model with attention modules smoothly goes down throughout the whole training process while we notice some sudden bumps around 60 to 80 epochs for ResNet32 without attention modules. 
To make sure that the results were not just an anomaly of ResNet32 on the particular Kaggle SAP dataset, we evaluated the performance of another architecture (ResNet26) with and without attention modules on a different dataset. Due to the scarcity of steering angle prediction datasets, the new dataset was created by the authors using the Udacity self-driving-car simulator \citep{Udacity2016}, which consists of approximately 10,765 images. We refer to this dataset as the custom dataset and publish it at \citep{barua2023}.
Figure \ref{fig:MSE-res26} shows a comparison between the mean squared error (MSE) scores of the two models during validation: the baseline ResNet26 model and the ResNet26 model with integrated attention modules. The MSE scores are plotted against the number of epochs. 
According to Figure \ref{fig:MSE-res26}, the MSE of the attention-aware ResNet26 model is significantly decreased with the number of epochs.
Except for epoch 70, where the performance of both models was similar, attention modules improved the performance of the ResNet26 baseline model. 
After 70 epochs, our model began to decay smoothly, while we could still see some fluctuations in the MSE score for the ResNet26 without attention modules. At epoch 100, we can observe a 6.83\% reduction in MSE for the ResNet32 on the Kaggle SAP dataset and a 6.09\% reduction for the ResNet26 on the custom dataset when the attention mechanism was employed. Figure \ref{MSE-decrease} illustrates the percentage reduction in MSE at 100 epochs during validation. These findings show that the integration of attention modules can yield clearly better steering angle prediction.

In addition to improved accuracy, we are also interested in our introduced attention mechanism's influence on adversarial robustness. 
To this end, we employed two well-established adversarial attack techniques: the Fast Gradient Sign Method (FGSM)\cite{Goodfellow2014} and Projected Gradient Descent (PGD) \citep{madry2017}.
As far as we know, we are the first to test the adversarial robustness of attention-equipped deep models for steering angle prediction.
The results of our adversarial attack experiments, including the MSE scores of ResNet32 and ResNet26 with and without integrated attention modules under FGSM and PGD attacks, are presented in Table~\ref{table:4} and Table~\ref{table:5}.
Epsilon (eps) values control the magnitude of adversarial perturbations.
While the MSE loss does increase under attacks, our results indicate that, with the inclusion of attention modules, the degree of this increase is notably mitigated. As we can see, attention boosts the model robustness by up to 53.95\% on ResNet32 (Table~\ref{table:4}) and as high as 56.09\% on ResNet26 (Table~\ref{table:5}).

\begin{table}
  \centering
  \renewcommand{\arraystretch}{1.2}
  \begin{tabular}{p{2cm}|c|c|c|c}
    \hline
    \multirow{2}{5cm}{\textbf{Model}} & \multicolumn{2}{c|}{\textbf{FGSM}} & \multicolumn{2}{c}{\textbf{PGD}}\\
    \cline{2-5}
    & eps=0.01 & eps=3 & eps=0.01 & eps=3\\
    \hline\hline
    w/o attention & 0.214 & 0.336 & 4.763 & 5.853 \\ \hline
    w attention & 0.200 & 0.291 & 2.581 & 2.695 \\ \hline
    change & 6.54\% & 13.39\% & 45.81\% & 53.95\% \\ \hline
  \end{tabular}
  \caption{Adversarial attack results on ResNet32 with and without attention. The value reported for each case is the MSE error of steering angle prediction. eps, or epsilon, controls the magnitude of the adversarial perturbation.}
  \label{table:4}
\end{table}

\begin{table}
  \centering
  \renewcommand{\arraystretch}{1.2}
  \begin{tabular}{p{1.5cm}|c|c|c|c}
    \hline
    \multirow{2}{5cm}{\textbf{Model}} & \multicolumn{2}{c|}{\textbf{FGSM}} & \multicolumn{2}{c}{\textbf{PGD}}\\
    \cline{2-5}
    & eps=0.5 & eps=0.7 & eps=0.5 & eps=0.7\\
    \hline\hline
    w/o attention & 0.253 & 0.574 & 9.464 & 9.636 \\ \hline
    w attention & 0.183 & 0.252 & 5.558 & 5.616 \\ \hline
    change & 27.66\% & 56.09\% & 41.27\% & 41.71\% \\ \hline
  \end{tabular}
  \caption{Adversarial attack results on ResNet26 with and without attention. The value reported for each case is the MSE error of steering angle prediction. eps, or epsilon, controls the magnitude of the adversarial perturbation.}
  \label{table:5}
\end{table}


     


\subsection{Qualitative Results} \label{sec:qualitative}
  

\begin{figure*}[!htb]
    \centering
    \begin{subfigure}{0.95\textwidth}
    \centering
            \includegraphics[width=.3\textwidth]{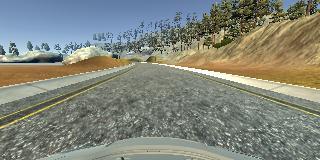}
            \includegraphics[width=.3\textwidth]{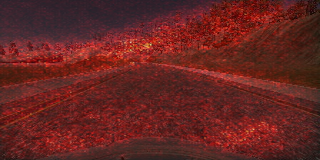}
            \includegraphics[width=.3\textwidth]{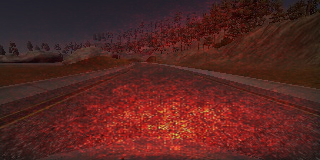}
    \end{subfigure}
    \hfill
    \begin{subfigure}{0.95\textwidth}
    \centering
            \includegraphics[width=.3\textwidth]{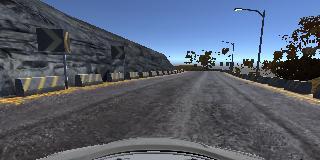}
            \includegraphics[width=.3\textwidth]{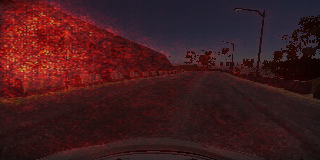}
            \includegraphics[width=.3\textwidth]{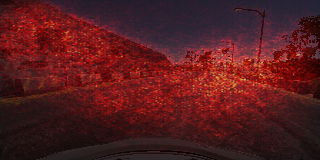}
   \end{subfigure}
   \hfill
    \begin{subfigure}{0.95\textwidth}
    \centering
            \includegraphics[width=.3\textwidth]{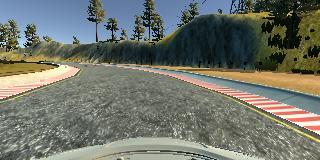}
            \includegraphics[width=.3\textwidth]{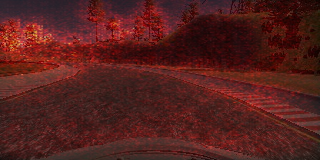}
            \includegraphics[width=.3\textwidth]{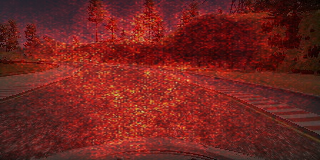}
   \end{subfigure}
   \hfill
    \begin{subfigure}{0.95\textwidth}
    \centering
            \includegraphics[width=.3\textwidth]{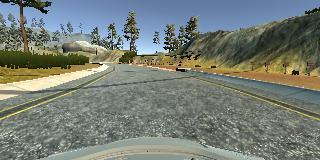}
            \includegraphics[width=.3\textwidth]{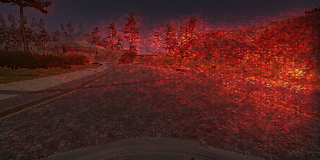}
            \includegraphics[width=.3\textwidth]{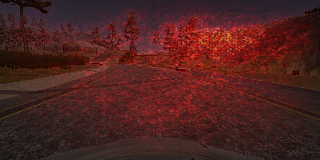}
            \caption*{(L) Original image, (M) ResNet32 w/o attention, (R) ResNet32 with attention}
    \end{subfigure}
    \caption{Saliency maps representing the focus areas of the model with and without our introduced attention modules. Black signifies no focus, red indicates some focus, and yellow indicates the highest level of focus. To generate a saliency map we get a predicted output from the model. We take the max score of this output and perform back-propagation on the max score with respect to the original input. We use layer 4 saliency as an example. For ResNet32 and ResNet32 with attention, the saliency maps are blended with the original image with a blending ratio of 0.75.}
    \label{fig:saliency}
\end{figure*}

To better understand the influence of our introduced attention modules, we illustrated the changes in saliency maps, which provide insight into where the model is focused on in an image (Fig. \ref{fig:saliency}). 
According to the results, we can see that the attention modules assist ResNet in directing its focus toward the road ahead, prioritizing it over irrelevant areas. Sometimes (Row 2 and Row 4 in Fig. \ref{fig:saliency}), without our attention mechanism, the original ResNet model focuses only on one side of the road, and with the attention mechanism added, the model can pay attention to the center and the other side of the road to a greater extent. Fig. \ref{fig:saliency} qualitatively explained our approach's efficacy.

\section{Conclusion}\label{section-universal}
Targeting the task of steering angle prediction, this paper first examines the two main veins of modern neural network architectures and their intra-variations of various complexity (i.e., InceptionNets and ResNets). 
In addition, we have introduced attention modules to deep steering angle predictors, which has increased the models’ performance by decreasing the MSE up to 6.83\% for the SAP dataset and up to 6.09\% for the custom dataset. 
The incorporation of our attention modules also demonstrates a promising capability to boost steering angle predictors' adversarial robustness. For example, it can enhance the adversarial robustness of ResNet32 by up to 53.96\% against the PGD attack.

\bibliographystyle{IEEEtran}
\bibliography{references.bib}

\end{document}